\documentclass[final]{cvpr}

\usepackage[linesnumbered,boxed,ruled,commentsnumbered]{algorithm2e}
\usepackage{times}
\usepackage{epsfig}
\usepackage{graphicx}
\usepackage{amsmath}
\usepackage{amssymb}
\usepackage[utf8]{inputenc} 
\usepackage[T1]{fontenc}    
\usepackage{url}            
\usepackage{booktabs}       
\usepackage{amsfonts}       
\usepackage{nicefrac}       
\usepackage{microtype}      
\usepackage{graphicx}
\usepackage{multirow}
\usepackage{wrapfig}
\usepackage{xcolor}
\usepackage{bbm}
\usepackage{amsmath}
\usepackage{comment}
\usepackage{dsfont}
\usepackage{bbm}
\DeclareUnicodeCharacter{2212}{-}
\usepackage[pagebackref=true,breaklinks=true,colorlinks,bookmarks=false]{hyperref}

\def\cvprPaperID{6084} 
\def\confYear{CVPR 2021}

\begin{document}

\title{One Thing One Click: A Self-Training Approach for\\ Weakly Supervised 3D Semantic Segmentation}

\author{Zhengzhe Liu$^{1}$ \quad   Xiaojuan Qi$^{2}$ \quad     Chi-Wing Fu$^{1}$ \\
$^1$The Chinese University of Hong Kong \quad $^2$The University of Hong Kong\\
{\tt\small \{zzliu,cwfu\}@cse.cuhk.edu.hk \quad  xjqi@eee.hku.edu.hk}
}

\maketitle

\begin{abstract}
Point cloud semantic segmentation often requires large-scale annotated training data, but clearly, point-wise labels are too tedious to prepare.
While some recent methods propose to train a 3D network with small percentages of point labels, we take the approach to an extreme and propose ``One Thing One Click,'' meaning that the annotator only needs to label one point per object.
%
To leverage these extremely sparse labels in network training, we design a novel self-training approach, in which we iteratively conduct the training and label propagation, facilitated by 
a graph propagation module.
Also, we adopt a relation network to generate the per-category prototype and explicitly model the similarity among graph nodes 
to generate pseudo labels to guide the iterative training.
%
Experimental results on both ScanNet-v2 and S3DIS show that our self-training approach, with extremely-sparse annotations, outperforms all existing weakly supervised methods for 3D semantic segmentation by a large margin, and our results are also comparable to those of the fully supervised counterparts.

\end{abstract}

\section{Introduction}

\begin{figure}
\centering
\includegraphics[width=0.99\columnwidth]{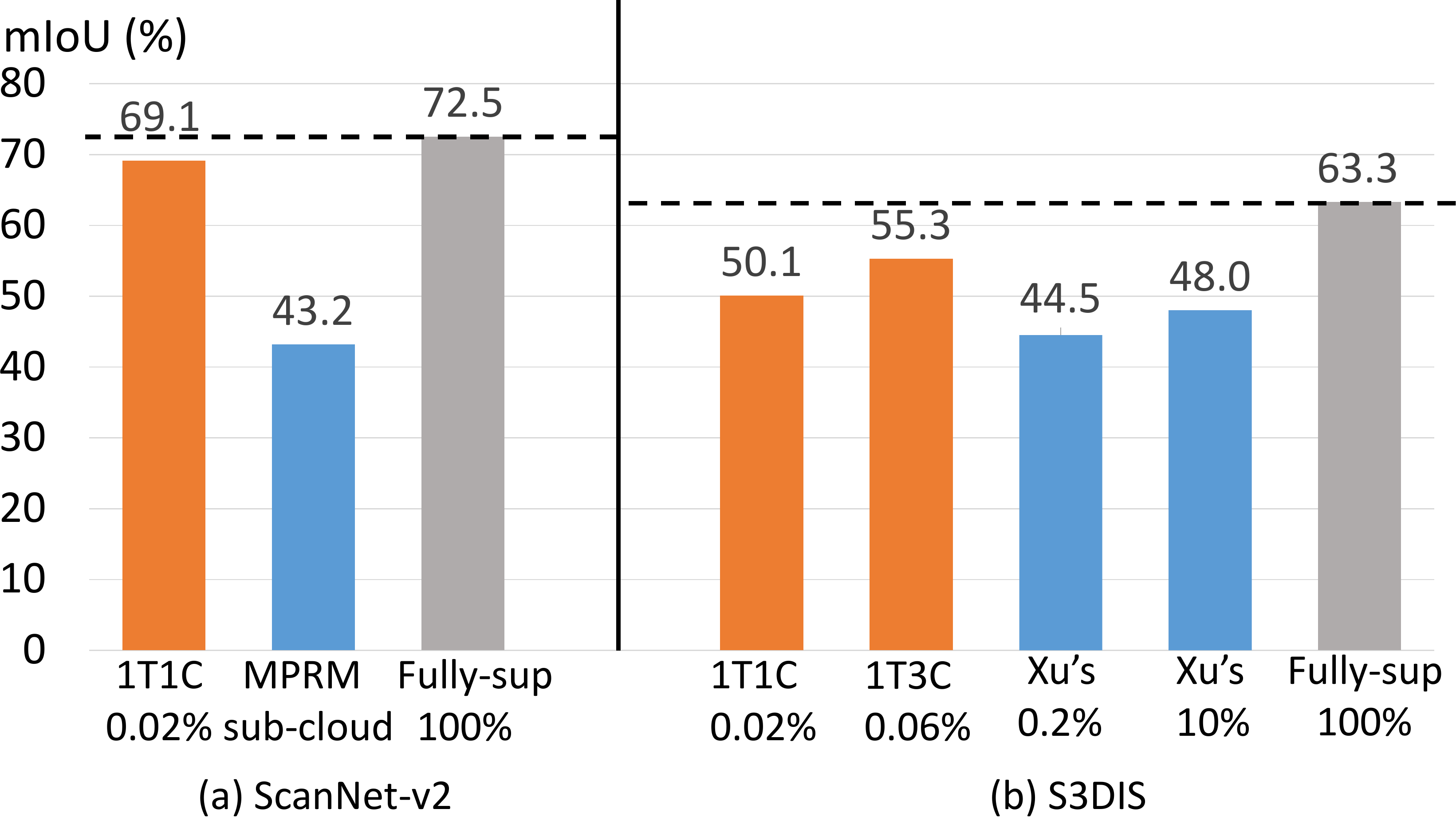}
\caption{Comparing our approach of ``One Thing One Click'' (1T1C) with two recent weakly supervised methods MPRM~\cite{wei2020multi} (CVPR 2020) and Xu's~\cite{xu2020weakly} (CVPR 2020) and a fully supervised version of our method Fully-sup on 3D semantic segmentation of ScanNet-v2 and S3DIS.
Our approach achieves better performance by {\em training on data with only one label per object\/}.
Note the annotation percentages under each method in the charts.
If ``One Thing Three Clicks'' (1T3C) is allowed, we can further raise our result.
}\label{fig:illustration}
 \vspace*{-1mm}
\end{figure}

\begin{figure}
\centering
\includegraphics[width=0.99\columnwidth]{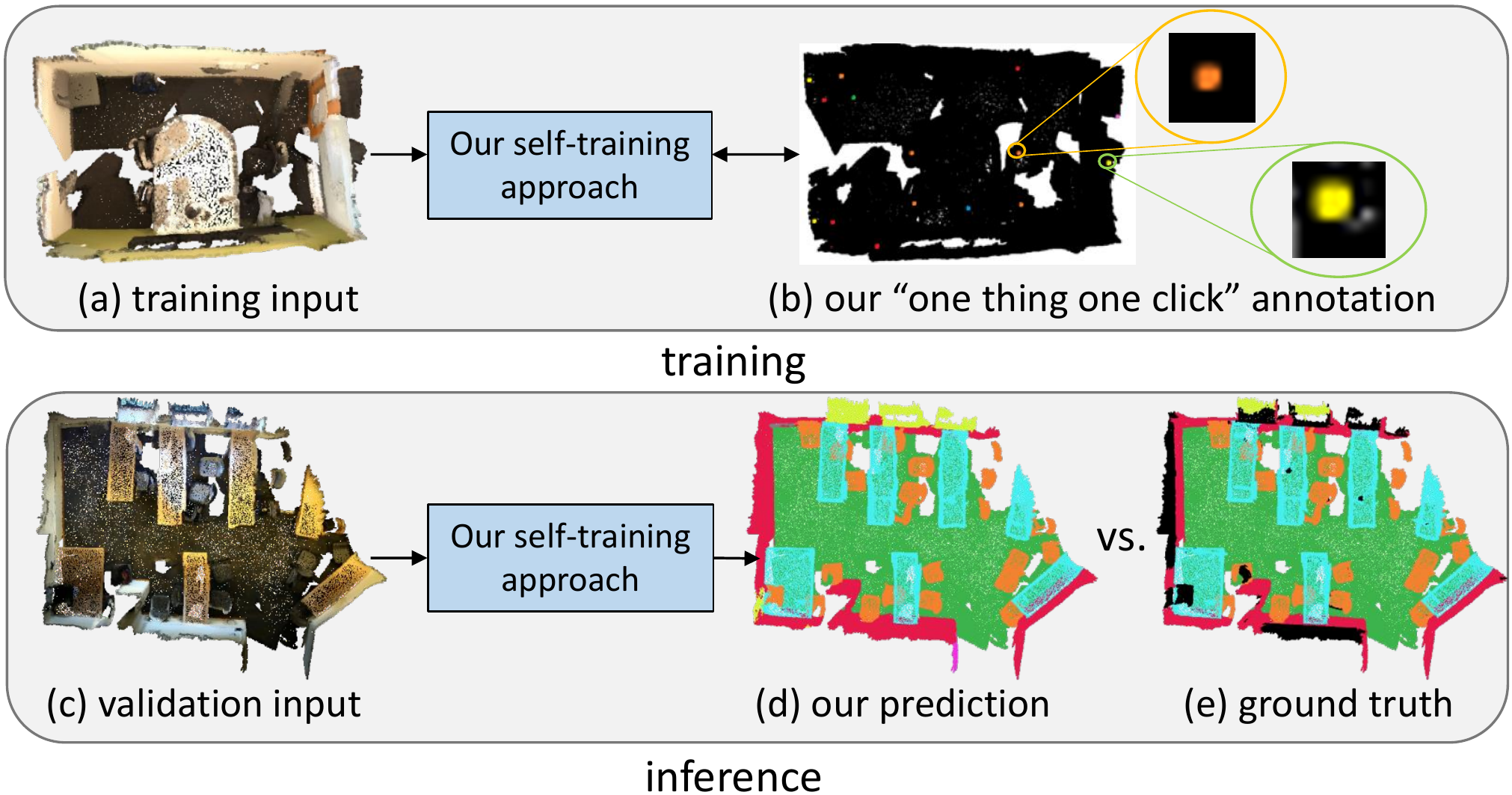}
\caption{We train our self-training approach using only our ``One Thing One Click'' annotations (top).
Yet, it can produce plausible segmentation results close to the ground truth (bottom).}
 \label{fig:fig2}
 \vspace*{-1mm}
\end{figure}


The success of 3D semantic segmentation benefits a lot from the large annotated training data. However, annotating a large amount of point cloud data is exhausting and costly. Taking ScanNet-v2\cite{dai2017scannet} as an example, it takes 22.3 minutes to annotate one scene on average. It is a great burden to annotate the whole data set, which includes 1,513 scenes, thus potentially restricting further applications that require larger scale data.
Thus, efficient approaches to facilitate 3D data annotation are highly desirable.

Very recently, some methods~\cite{wei2020multi,wang2020weakly,xu2020weakly} were proposed to reduce efforts to annotating 3D point clouds.
Though they improve annotation efficiency, various issues remain.
Scene-level annotation in~\cite{wei2020multi} could impose negative effects on the model in the absence of localization information, whereas sub-cloud annotation in~\cite{wei2020multi} requires an extra burden to first divide the input into subclouds and then repeatedly annotate semantic categories in individual subclouds.
The 2D image annotation approach~\cite{wang2020weakly} requires extra labor to prepare a 2D image annotation, which is also a tedious task on its own.
Xu~\etal~\cite{xu2020weakly} presume that the labeled points follow a uniform distribution.
Such a requirement can be achieved by subsampling from a fully-annotated dataset, but is hard for the annotators to follow in practice.

In this work, we also aim to reduce the amount of necessary annotations on point clouds, but we take the approach to an extreme by proposing ``One Thing One Click,'' so the annotator only needs to label one single point per object.
To further relieve the annotation burden, such a point can be randomly chosen, not necessarily at the center of the object.
On average, it takes less than 2 minutes to annotate a ScanNet-v2 scene with our ``One Thing One Click'' scheme (see an example annotation in Figure~\ref{fig:fig2} (b), which contains only 13 clicks), which is more than 10x faster compared with the original ScanNet-v2 annotation scheme.

However, directly training a network on the extremely-sparse labels from our annotating scheme (less than 0.02\% in ScanNet-v2 and S3DIS) will easily make the network overfit the limited data and restrict its generalization ability.
Hence, it raises a question that ``can we achieve a performance comparable with a fully supervised baseline given the extremely-sparse annotations?''
To meet such a challenge, we design a self-training approach with a label-propagation mechanism for weakly supervised semantic segmentation.
On the one hand, with the prediction result of the model, the pseudo labels can be expanded to unknown regions through our graph propagation module.
On the other hand, with richer and higher quality labels being generated, the model performance can be further improved.
Thus, we conduct the label propagation and network training iteratively, forming a closed loop to boost the performance of each other.


A core problem of label propagation is how to measure the similarity among nodes. Previous works~\cite{zheng2015conditional,chen2017deeplab,yuan2019structpool} build a graph model upon 2D pixels and measure the similarity with low-level image features, e.g., coordinates and colors.
In contrast, our graph is built upon the 3D super-voxels with more complex geometric structures and a variable number of points in each group. Hence, existing hand-craft features cannot fully reveal the similarity among nodes in our case.
To resolve this problem, we further propose a relation network to leverage 3D geometrical information for similarity learning among the graph nodes in 3D. The geometrical similarity and learned similarity are integrated together to facilitate label propagation. To effectively train the relation network with the extremely-sparse and category-unbalanced data, we further propose to generate a category-wise prototype with a memory bank for better similarity measurement.


Experiments conducted on two public data sets ScanNet-v2 and S3DIS manifest the effectiveness of the proposed method.
With just around 0.02\% point annotations, our approach surpasses all existing weakly supervised approaches (which employ far more labels) for 3D point cloud segmentation by a large margin, and our approach even achieves results that are comparable with a fully supervised counterpart; see Figure~\ref{fig:illustration}. These results manifest the high efficiency of our ``One Thing One Click'' scheme for 3D point cloud annotation and the effectiveness of our self-training approach for weakly supervised 3D semantic segmentation.

\begin{figure*}
\centering
\includegraphics[width=0.99\textwidth]{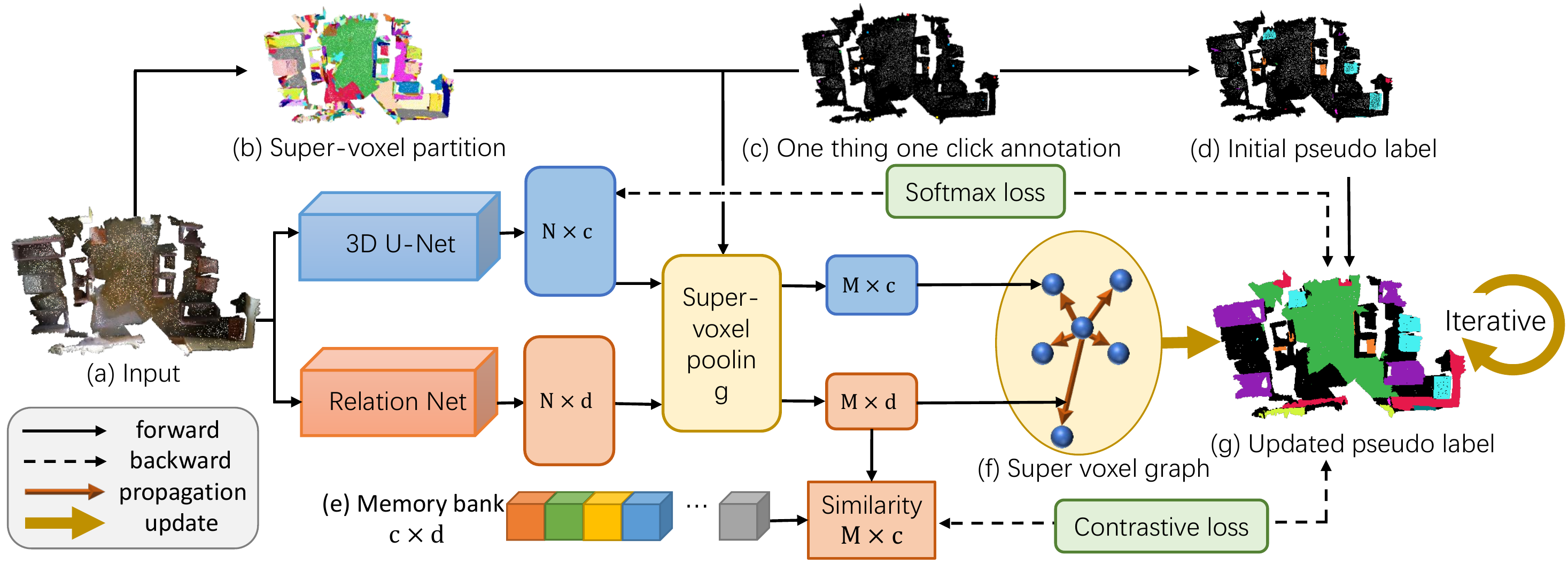}
\caption{Overview of our framework.
Through a super-voxel partition (b), we expand our ``One Thing One Click'' annotations (c) to generate the initial pseudo labels (d) for guiding the update of the pseudo labels (g).
On the other hand, we adopt the ``3D U-Net'' for semantic label prediction (blue region) and design the ``Relation Net'' for super-voxel-based similarity learning (orange region).
Then, we incorporate a super-voxel pooling to aggregate features from the two networks and construct the super-voxel graph (f) to propagate labels over the point cloud.
Further, we iteratively update the predicted labels (g) and train the network through the softmax loss and contractive loss.
$C$ is the number of categories, $D$ is the number of the feature dimension, $N$ is the number of points, and $M$ is the number of super-voxels.}
\label{fig:overview}
\vspace*{-2mm}
\end{figure*}

\section{Related Work}

\paragraph{Semantic Segmentation for Point Cloud}
Approaches for 3D semantic segmentation can be roughly divided into point-based methods and voxel-based methods.
\textit{Point-based networks} take raw point clouds as input.  Along this line of works, PointNet~\cite{qi2017pointnet} and PointNet++~\cite{qi2017pointnet++} are the pioneering ones. 
Afterward, convolution-based methods~\cite{li2018pointcnn,thomas2019kpconv,wu2019pointconv,boulch2020convpoint} were also proposed for 3D semantic segmentation on point clouds. Recently, Kundu~\etal~\cite{kundu2020virtual} proposed to fuse features from multiple 2D views for 3D semantic segmentation.
To aggregate together the geometrically-homogeneous points, Landrieu\etal~\cite{landrieu2018large} modeled a point cloud as a super point graph. Inspired by~\cite{landrieu2018large}, we expand the sparse labels to geometrically homogeneous super-voxels to generate initial pseudo labels for the first-iteration network training.

\textit{Voxel-based networks} take the regular voxel-grids as input instead of the raw data ~\cite{tchapmi2017segcloud,riegler2017octnet,graham2015sparse,su2018splatnet,dai20183dmv}. 
The recently-proposed methods SparseConv~\cite{graham2017submanifold}, MinkowskiNet~\etal~\cite{choy20194d}, and OccuSeg~\etal~\cite{han2020occuseg} are among the representative works in this branch. 
In this paper, we adopt the 3D-UNet architecture described in~\cite{graham2017submanifold} as the backbone architecture due to its high performance and applicability.

\vspace*{-2.5mm}
\paragraph{Weakly Supervised 3D Semantic Segmentation} Compared with fully supervised 3D semantic segmentation, weakly supervised 3D semantic segmentation is relatively under-explored.
After early works~\cite{mei2019semantic,guinard2017weakly} in this area, very recently, 
Wei~\etal~\cite{wei2020multi} utilized the Class Activation Map to generate pseudo point-wise labels from sub-cloud-level annotations.
The performance is, however, limited by the lack of localization information.
Wang~\etal~\cite{wang2020weakly} back-projected 2D image annotations to 3D space to produce labels in point clouds.
However, annotating large-scale semantic segmentation on 2D images is also laborious.
Also, the visibility prediction branch adds to the complexity of the network.  
Xu~\etal~\cite{xu2020weakly} achieve a performance close to fully supervised with less than 10\% labels.
However, they require the annotations to be uniformly-distributed in the point cloud, which is practically very hard for the annotators to follow. 

Different from the existing works, we propose a new self-training approach with a graph propagation module, in which the network training and label propagation are conducted iteratively.
Our approach largely reduces the reliance on the quality of the initial annotation and achieves top performances, compared with existing weakly supervised methods, while using only extremely-sparse annotations.



\vspace*{-2.5mm}
\paragraph{Self-Training}
Self-training for weakly supervised 2D image understanding has been intensively explored.
To reduce the annotation burden for 2D images, researchers proposed a variety of annotation approaches,~\eg, image-level categories~\cite{qi2016augmented,oh2017exploiting,zhou2018weakly,ahn2018learning}, points~\cite{bearman2016s,laradji2018blobs}, extreme points~\cite{maninis2018deep,papadopoulos2017extreme},  scribbles~\cite{lin2016scribblesup,wang2019boundary,zhang2020weakly}, bounding boxes~\cite{dai2015boxsup}, etc. With the weak supervision, a self-training approach can learn to expand the limited annotations to unknown regions in the domain. As far as we know, this is the first work that explores self training for weakly supervised 3D semantic segmentation.

\section{Methodology}

\subsection{Overview}

With ``One Thing One Click,'' we only need to annotate a point cloud with one point per object, as Figure~\ref{fig:overview} (c) shows, and these points can be chosen at random to alleviate the annotation burden.  
Procedure-wise, given such sparse annotations, we first over-segment the point cloud $X=\{p_i\}$ into geometrically homogeneous super-voxels $V=\{v_j\}$, where $\cup_j v_j=X$ and $v_j \cap v_{j'}=\emptyset$ for $v_j \neq v_{j'}$.
Note that throughout the paper, we use $i$ and $j$ as the indices for points and super-voxels, respectively.
Based on the super-voxel partition, we can produce initial pseudo labels of the point cloud by spreading each label to all the points locally in the super-voxel that contains the annotated point.
However, as Figure~\ref{fig:overview} (d) shows, the labels are still very sparse. More importantly, the propagated labels distribute mainly around the initially-annotated points, which are far from the ideal uniform distribution for weakly semantic segmentation, as employed in ~\cite{xu2020weakly}.


An important insight in our approach is to iteratively propagate the sparse annotations to unknown regions in the point cloud, while training the network model to guide the propagation process.  To achieve this, we adopt the 3D semantic segmentation network $\Theta$ (the blue regions in Figure~\ref{fig:overview}) to learn to propagate via a graph model (Figure~\ref{fig:overview} (f)). Further, we design the relation network $\mathcal{R}$ (the orange regions in Figure~\ref{fig:overview}) to explicitly model the feature similarity among the graph nodes. Afterward, predictions with high confidence are further employed as the updated pseudo labels for training the network in the next iteration (Figure~\ref{fig:overview} (g)).
This iterative self-training approach couples the label propagation and network training, enabling us to significantly enhance the segmentation quality, as revealed earlier in Figure~\ref{fig:illustration}.


In this section, we first present our 3D semantic segmentation network for point-wise semantic prediction (Section~\ref{sec:unary}), then our label propagation mechanism with a graph model and the relation network for similarity learning (Section~\ref{sec:pairwise}). Afterward, we describe the self-training approach that evolves the above modules alternatively (Section~\ref{sec:self-training}).

\subsection{3D Semantic Segmentation Network}\label{sec:unary}
We adopt the 3D U-Net architecture~\cite{graham2017submanifold} as the backbone, denoted as $\Theta$. 
Its input is point cloud $X$ of $N$ points (Figure~\ref{fig:overview} (a)).
Each point has 3D coordinates $p_i$ and color $c_i$, where $i\in\{1, ..., N\}$. The network predicts the probability of each semantic category $P(y_{i,\bar{c}}|p_i,c_i,\Theta)$ of each point $p_i$, where $\bar{c}$ is the ground truth category of point $p_i$. The network is trained with the softmax cross-entropy loss below:
%
\vspace*{-1.5mm}

\begin{equation}
L_{s} = −\frac{1}{N}\sum^N_{i=1}{\log P({y_{i,\bar{c}}|p_i,c_i,\Theta})}
\label{equ:seg_loss}
\end{equation}

\vspace*{-1mm}
\noindent
In the first iteration, the network is trained with the initial pseudo labels, as shown in Figure~\ref{fig:overview} (d).
In subsequent iterations, the network is trained with the updated pseudo labels, as shown in Figure~\ref{fig:overview} (g), which will be detailed below.

\subsection{Pseudo Label Generation by Graph Propagation}\label{sec:pairwise}

To facilitate the network training, we propose a graph propagation mechanism to effectively propagate labels to unknown regions. We also propose the relation network to explicitly learn the similarity among the super-voxels to facilitate the label propagation process and complement 3D U-Net. 

\vspace*{-2.5mm}
\paragraph{Graph Construction}


To start, we leverage the 3D geometrically homogeneous super-voxels to build a graph.
Compared with building on points, our graph has significant fewer nodes to facilitate efficient label propagation.

To derive the prediction $P(y_{j,c}|v_j,\Theta)$ of the $j$-th super-voxel,
we apply a super-voxel pooling to aggregate the semantic prediction of the $n_j$ points in $v_j$ as below:
%
\begin{equation}
\label{equ:average}
P(y_{j,c}|v_j,\Theta) = \frac{1}{n_j} \sum_{i} P({y_{i,c}|p_i,c_i,\Theta}), \ \text{where} \ p_i \in v_j,
\end{equation}
where 
$P({y_{i,c}|p_i,c_i,\Theta})$ is the probability of $p_i$ in class $c$.

To build the graph, we treat each super-voxel as a graph node and compute the similarity between each pair of super-voxels $v_j,v_{j'}$, which is represented as an edge. Further, to propagate labels to unknown regions through the graph, we formulate it as an optimization problem that considers both the network prediction and similarities among the super-voxels to achieve the global optimum with the energy function below similar to Conditional Random Field (CRF).
\begin{equation}
\label{equ:energy}
E(Y|V) = \sum_{j} \psi_u (y_{j}|V,\Theta) + \sum_{j<{j'}} \psi_p (y_{j},y_{j'} | V,\mathcal{R},\Theta )
\end{equation}
where $\mathcal{R}$ is the relation network to be detailed later.
The unary term $\psi_u (y_{j}|V,\Theta)$ represents the super-voxel pooled prediction of the 3D U-Net $P(y_j)$ on super-voxel $v_j$. Specifically, it denotes the minus $\log$ probability of predicting super-voxel $v_j$ to have label $y_{j}$. 
We define it as below. 
\begin{equation}
\label{equ:unary_loss}
 \psi_u (y_{j}|V,\Theta) = -\log P({y_{j}|V,\Theta})
\end{equation}

The pairwise term $\psi_p (j_k)$ in Equation~\ref{equ:energy} represents the similarity between super-voxels $v_j$ and $v_{j'}$. We employ both the low-level features and learned features for measuring the similarity, as shown in Equation~\ref{equ:pairwise} below:
\vspace*{-1.5mm}
\begin{equation}
\begin{aligned}
\label{equ:pairwise}
\psi_p (y_j,y_{j'} | V)= & \mathds{1} (y_j , y_{j'}) \exp\{- \lambda_c \frac{ \left\Vert c_j-c_{j'} \right\Vert^2}{2\sigma^2_c} \\
- \lambda_p \frac{  \left\Vert p_j-p_{j'} \right\Vert^2}{2\sigma^2_p} 
& - \lambda_u \frac{ \left\Vert u_j-u_{j'} \right\Vert^2}{2\sigma^2_u} 
- \lambda_f \frac{ \left\Vert f_j-f_{j'} \right\Vert^2}{2\sigma^2_f} 
\}
\end{aligned}
\end{equation}
where $\mathds{1} (y_j, y_{j'})$ is 1, if $v_j$ and $v_{j'}$ have different predicted labels, and 0 otherwise. The pairwise term means that the cost will be higher if super-voxels with similar features are predicted to be different classes.
Here, $c_j,c_{j'}$, $p_j,p_{j'}$ and $u_j,u_{j'}$ are the normalized mean color, mean coordinates and mean 3D U-Net feature, respectively, of super-voxels $v_j$ and $v_{j'}$.
Unlike existing works~\cite{zheng2015conditional,chen2017deeplab,yuan2019structpool}, which build the graph on 2D image pixels, we build our graph on 3D super-voxels, which have irregular and complex geometrical structures, as shown in the supplementary material. Therefore, hand-crafted features $p_j,p_{j'}$ and $c_j,c_{j'}$ have inferior capability to measure the similarity between super-voxels. To address this issue, we propose the \textit{Relation Network} to better leverage the 3D geometrical information and explicitly learn the similarity among super-voxels.


\vspace*{-2.5mm}
\paragraph{Relation Network}

Existing works Co-Training~\cite{qiao2018deep} and Tri-net~\cite{dong2018tri} showed that semi-supervised training benefits from having two complementary tasks or components. In our framework, we propose a relation net to complement the 3D U-Net.
The relation network $\mathcal{R}$ shares the same backbone architecture as the 3D U-Net $\Theta$ except for removing the last category-wise prediction layer. It aims to predict a category-related embedding $f_j$ for each super-voxel $v_j$ as the similarity measurement. Similar to Equation~\ref{equ:average}, $f_j$ is the per super-voxel pooled feature in $\mathcal{R}$. In other words, the relation network groups the embeddings of same category together, while pushing those of different categories apart. To this end, we propose to learn a prototypical embedding for each category, inspired by the Prototypical Network~\cite{snell2017prototypical}.


However, the per-category prototypes in~\cite{snell2017prototypical} are fully determined by the sampled mini-batch, and may deviate from the actual categorical center. Consequently, they may not be stable and could keep changing during the training, thereby hard to converge. To assist the training of the relation network with sparse and unbalanced training data, we present a memory bank $K=\{k\}$ to generate one categorical prototype for each category, instead of simply regarding the average embedding as the prototype as in~\cite{snell2017prototypical}. 

The embedding $f_j$ generated by $\mathcal{R}$ serves as a ``query,'' and we compare it with the corresponding ``key'' $k_c$ in the memory bank with a dot product.
The two modules are optimized simultaneously with contrastive learning~\cite{oord2018representation} as below.

\vspace*{-1.5mm}
\begin{equation}
\label{equ:contrastive}
L_{c} =\frac{1}{M}\sum^{M}_{j} {(-\log{   \frac{f_j \cdot k_{\bar{c}}/\tau}{\sum_c f_j \cdot k_c/\tau}         })},
\end{equation}

\vspace*{-2mm}
\noindent
where $\tau$ is a temperature hyperparameter~\cite{wu2018unsupervised} and $\bar{c}$ is the ground truth category of $v_j$. The contrastive learning is equivalent to a c-way softmax classification task. 

Following~\cite{he2020momentum}, we update the key representations via a moving average with momentum as shown below
\begin{equation}
\label{equ:momentum}
k_{\bar{c}} \xleftarrow{} m k_{\bar{c}} +(1-m) f_j,
\end{equation}
where $m$ is a momentum coefficient to control the evolving speed of the memory bank. On the one hand, the representations in the memory bank are initialized with random vectors, and are updated during training to generate the prototype for each category. On the other hand, the embeddings generated from the relation network are grouped towards the prototype of its category. In this way, the relation network generates similar embeddings for the same category and distinct ones for different categories. The memory bank updates the prototypes in a category-balanced manner by randomly sampling the same number of training samples $s$ per category in every forward pass. 

Our relation net complements with 3D U-Net. It measures the relations between super-voxels using different training strategies and losses, while 3D U-Net aims to project the inputs into the latent feature space for category assignment. The prediction of relation network is further combined with the prediction of 3D U-Net by multiplying the predicted possibilities of each category to boost the performance. In addition, the relation net offers a stronger measurement of the pairwise term in CRF vs. handcrafted features like colors and also complements with the 3D U-Net features.

\subsection{Self-Training}\label{sec:self-training}

With the energy function in Equation~\ref{equ:energy}, we propose a self-training approach to update networks $\Theta$ and $\mathcal{R}$, and also the pseudo labels $Y$ iteratively, as Algorithm~\ref{selftrain} outlines. The self-training is started by the ``One Thing One Click'' annotations and the pre-constructed super-voxel graph. In each iteration, we fix network parameters $\Theta,\mathcal{R}$ and update label $Y$, and vice versa. There are two steps in each iteration.
\begin{itemize}
\vspace*{-1mm}
\item
With $\Theta$ and $\mathcal{R}$ fixed,
the label propagation is conducted to minimize the energy function in Equation~\ref{equ:energy}.  Then, the predictions with high confidence are taken as the updated pseudo labels for training the two networks in the next iteration.
The confidence of super-voxel $v_j$, denoted as $C_j$, is the average of the minus $\log$ probability of all $n_j$ points in $v_j$ after the label propagation:  
\vspace*{-1.5mm}
\begin{equation}
\label{equ:momentum}
C_j = \frac{1}{n_j} \sum_i^{n_j}{ \log P(y_i| p_i, V, \Theta, \mathcal{R}, G)}, \ \text{where} \ p_i \in v_j,
\end{equation}

\vspace*{-2.5mm}
\noindent
where $G$ denotes the graph propagation. 
\vspace*{-1mm}
\item
With pseudo labels $Y$, $\Theta$ and $\mathcal{R}$ are optimized with softmax loss and contrastive loss, respectively. 
\end{itemize}

\vspace*{-0.2in}
\section{Experiments}

\paragraph{Datasets}

Our experiments are conducted on two large 3D semantic segmentation datasets -- ScanNet-v2~\cite{dai2017scannet} and S3DIS~\cite{armeni2017joint}.  \textbf{ScanNet-v2}~\cite{dai2017scannet} contains $1513$ 3D scans of 20 semantic categories. We annotate the official training set with our ``One Thing One Click'' scheme, and evaluate on the original validation and test set. 
\textbf{S3DIS}~\cite{armeni2017joint} contains 3D scans of $271$ rooms containing $13$ categories. We follow the official train/validation split to annotate on Area 1,2,3,4,6 and report the performance on Area 5. 
\vspace*{-0.15in}
\paragraph{``One Thing One Click'' Annotation Details}
In order to ensure the randomness of point selection in annotation, we simulate the annotation procedure by selecting a single point inside an object with the same probability for the following experiments.
In ScanNet-v2, only 19.74 points per scene are annotated on average with ``One Thing One Click'' scheme, 
while this number in the original ScanNet-v2 is 108875.9. In S3DIS, only 36.15 points in each room are annotated on average using ``One Thing One Click'', while the original S3DIS has 193797.1 points annotated in each room. 
\vspace*{-0.15in}
\paragraph{Implementation Details}
We implement all the modules of our self-training framework including the mean-field solver~\cite{koller2009probabilistic} for label propagation with the PyTorch~\cite{NEURIPS2019_9015} framework based on the implementation of~\cite{jiang2020pointgroup}. 
Following~\cite{jiang2020pointgroup}, due to the GPU capacity, we randomly choose 250k points if the scene contains more points in training. In inference, the network takes the whole scene as input. 
We use the mesh segment results~\cite{dai2017scannet} as super-voxels for ScanNet-v2, and the geometrical partition results described in~\cite{landrieu2018large} for S3DIS super-voxel partition. We set the hyper-parameters $D=32$, $T=0.9$, $s=20$, $\tau=0.07$, $m=0.9$,
$\sigma_c=\sigma_p=\sigma_u=\sigma_f=1$,  $\lambda_c=\lambda_p=\lambda_u=\lambda_f=1$ with a small validation set.  
We found that the self-training converges after five iterations. After that, more iterations training only brings very minor improvements.

\IncMargin{1em}
\begin{algorithm}[!t]
    \SetAlgoNoLine 
    \SetKwInOut{Input}{\textbf{Input}}\SetKwInOut{Output}{\textbf{Output}} 
    \Input{``One Thing One Click'' annotations $Y_0=\{p_i\}$;\\
        super-voxel partition $V=\{v_j\}$\;\\}
    \Output{
        semantic prediction for all points Y$\;$\\}
    \BlankLine
    Expand the annotated points $p_i$ to the super-voxel $v_j$ if $p_i$ $\in$ $v_j$; \\
    \Repeat
        {\text{convergence}}
        {Train 3D U-Net $\Theta$ with pseudo labels $Y_t$; \\
        Train relation network $\mathcal{R}$ with pseudo labels $Y_t$; \\
        Combine the predictions and propagate the label with the graph model; \\
        Update the pseudo labels $Y_{t}$ to $Y_{t+1}$ with the predictions of high confidence. 
        }
    \caption{Our self-training approach.\label{selftrain}}
\end{algorithm}
\DecMargin{1em}


\subsection{Evaluations on ScanNet-v2}\label{sec:scannet}

\paragraph{Comparing with Existing Methods}

Table~\ref{tab:existing} reports the benchmark result on ScanNet-v2 test set. 
The baselines can be roughly divided into two branches.
(i) Fully supervised approaches with 100\% supervision, including several representative works in 3D semantic segmentation.
These methods are the upper bounds of weakly supervised ones.
(ii) Weakly supervised approaches, including a recent work~\cite{wei2020multi}.

With less than 0.02\% annotated points, our result (69.1\% mIoU) outperforms many existing works with full supervision. 
As for weakly supervised approaches, MPRM~\cite{wei2020multi} is trained with scene-level or subcloud-level labels. The scene-level annotation leads to an inferior performance of 24.4\%, and the subcloud-level annotation takes around 3 minutes per scene as reported in~\cite{wei2020multi}, which is longer than our ``One Thing One Click'' scheme (2 minutes). More importantly, our result outperforms~\cite{wei2020multi} by more than 26\% mIoU.

\begin{table}
\centering
\scalebox{0.85}{
  \begin{tabular}{c|cc}
    \toprule
    Method & Supervision & mIoU (\%)  \\
    \midrule
    Pointnet++~\cite{qi2017pointnet++} & 100\% &33.9 \\
    SPLATNet~\cite{su2018splatnet}& 100\% & 39.3\\
    TangentConv~\cite{tatarchenko2018tangent} & 100\% &43.8\\
    PointCNN~\cite{li2018pointcnn} & 100\% & 45.8\\
    FPConv~\cite{lin2020fpconv} & 100\% & 63.9\\
    DCM-Net~\cite{schult2020dualconvmesh}&100\%& 65.8 \\
    PointConv~\cite{wu2019pointconv} &100\%& 66.6 \\
    KPConv~\cite{thomas2019kpconv} & 100\% &68.4\\
    JSENet~\cite{hu2020jsenet}& 100\% &69.9 \\
    SubSparseCNN~\cite{graham2017submanifold} & 100\% & 72.5\\
    MinkowskiNet~\cite{choy20194d} &100\% & 73.6 \\
    Virtual MVFusion~\cite{kundu2020virtual} &100\%+2D & 74.6\\
    \midrule
    Our fully-sup baseline & 100\% & 72.5 \\
    \midrule
    MPRM~\cite{wei2020multi} & scene-level & 24.4 \\    
    MPRM~\cite{wei2020multi} & subcloud-level & 41.1 \\    
    MPRM+CRF~\cite{wei2020multi} & subcloud-level & 43.2 \\  
    One Thing One Click & 0.02\% & \textbf{69.1}\\ 
    \midrule
    CSC\_LA\_SEM~\cite{hou2020efficient} & 20 points/scene & 53.1 \\
    PointContrast\_LA\_SEM~\cite{xie2020pointcontrast} & 20 points/scene & 55.0 \\
    Ours on ``Data Efficient'' & 20 points/scene & \textbf{59.4}\\ 
    \bottomrule
  \end{tabular}
}
\caption{Comparing with existing methods and baselines on ScanNet-v2 test set.}
\label{tab:existing}
\end{table}

\begin{figure*}
\centering
\includegraphics[width=0.9\textwidth]{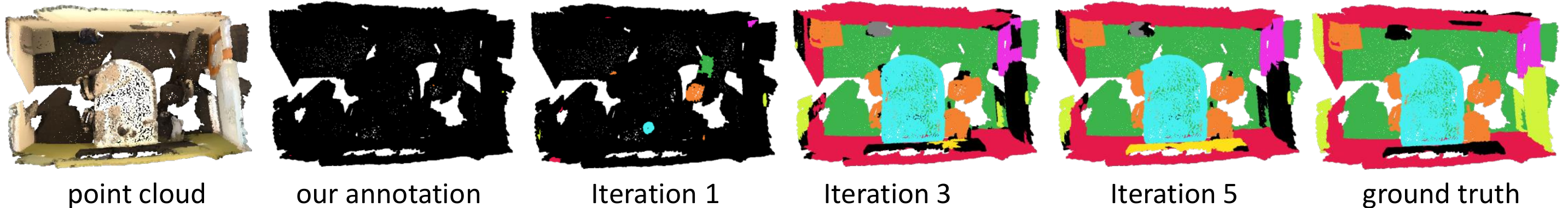}
\caption{Pseudo labels for each iteration on ScanNet-v2 training set. } \label{fig:iter}
\end{figure*}

\begin{figure*}
\centering
\includegraphics[width=0.9\textwidth]{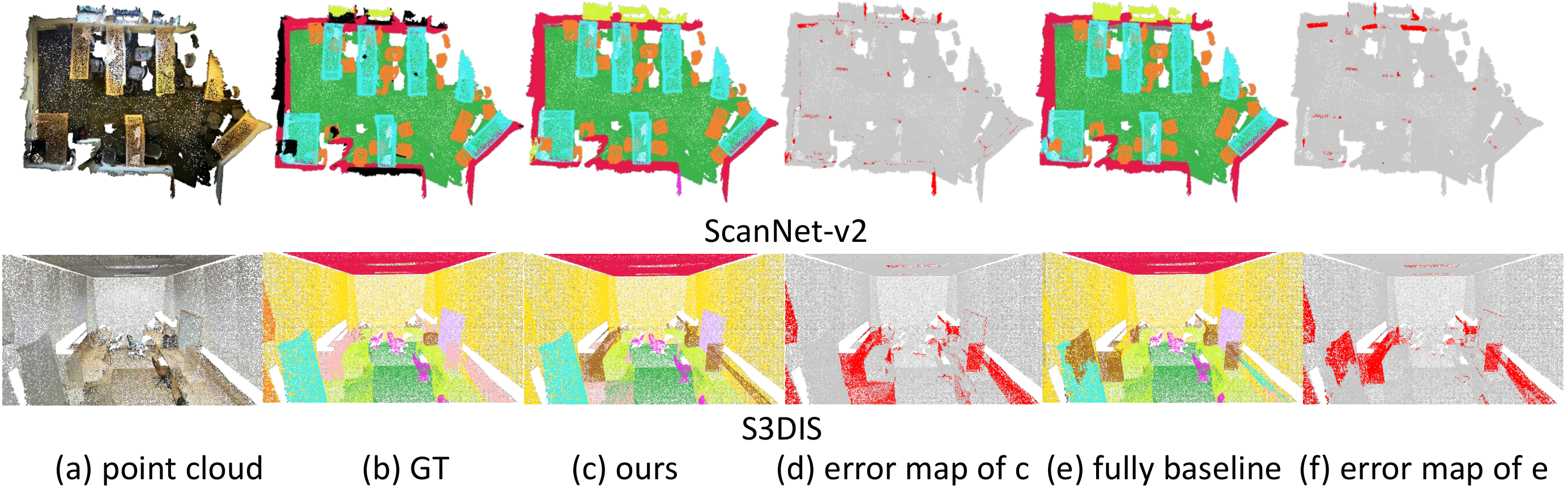}
\caption{Quantitative results of our method and fully supervised baseline. (d) is the error map of our prediction (c), and (f) is the error map of our fully supervised baseline~\cite{graham2017submanifold} (e). Red regions indicate the wrong prediction. } \label{fig:result}
\end{figure*}

\begin{figure}
\centering
\includegraphics[width=0.99\columnwidth]{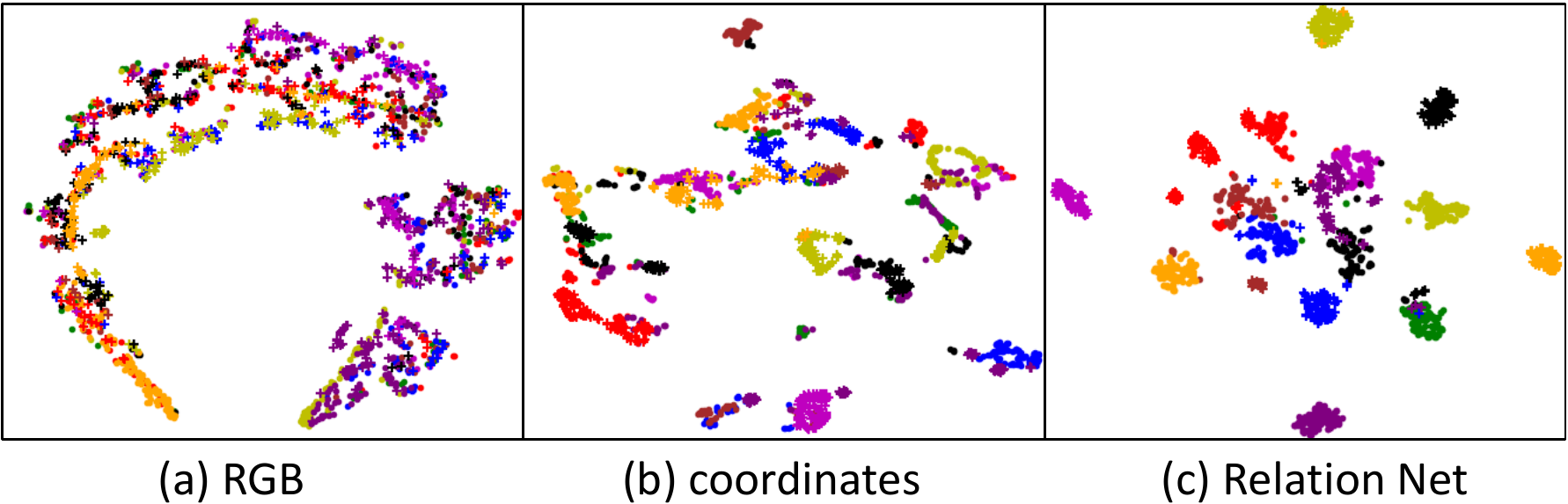}
\caption{The t-SNE visualization of super-voxel features. Different colors and marks (point and plus) indicate different categories. 
The samples of the same category are better grouped together with our relation network (c), compared with hand-crafted features (a \& b). } \label{fig:tsne}
\end{figure}

\vspace*{-0.1in}
\paragraph{Comparing with Our Baselines}

In this section, we first present three important baselines as shown in Table~\ref{tab:baseline} on ScanNet-v2 validation set. 
\begin{itemize}
\item Table~\ref{tab:baseline} ``Our fully sup baseline'' is trained with the official 100\% annotation provided by ScanNet-v2. It serves as the upper bound of our method. 
\item The model directly trained with the raw annotated points as Figure~\ref{fig:overview} (c) cannot converge well due to the extreme sparsity of the training data.
\item Table~\ref{tab:baseline} ``One Thing One Click$^*$''. The model trained with the initial pseudo labels as Figure~\ref{fig:overview} (d) achieves 62.18\% mIoU. It serves as the starting point of our self-training approach and is denoted as ``our baseline'' in the following. 
\end{itemize}

Table~\ref{tab:baseline} ``One Thing One Click'' manifests that our self-training approach surpasses the baseline by nearly $10\%$ mIoU, attaining a $16\%$ relative improvement. Compared with the fully supervised baseline with the same network architecture, our performance is only 2\% lower. 

Table~\ref{tab:baseline} ``One Thing One Click$^\dagger$'' refers to disabling the graph propagation and relation network in inference. Note that they are still being used in training for generating the pseudo labels. This brings no extra computational burden during the inference, but helps to improve nearly 7\% mIoU, comparing with the baseline (68.96\% vs 62.18\%).

The quantitative results in Figure~\ref{fig:result} indicate our result (c) is very similar to the fully supervised baseline (e)~\cite{graham2017submanifold} in ScanNet-v2. Check error maps (d) (f) for better comparison.

\begin{table}
\centering
\scalebox{0.9}{
  \begin{tabular}{c|ccc}
    \toprule
    Setting & Annotation & mIoU (\%) \\
    \midrule
    Our fully sup baseline & 100\% & 72.18 \\
    \midrule
    One Thing One Click$^*$ & 0.02\% & 62.18\\
    One Thing One Click$^\dagger$ & 0.02\% & 68.96\\
    One Thing One Click & 0.02\% & \textbf{70.45}\\
    \midrule
    Data Efficient$^*$ & 20 points & 55.06\\
    Data Efficient$^\dagger$ & 20 points & 59.98\\ 
    Data Efficient & 20 points & \textbf{61.35}\\ 
    \bottomrule
  \end{tabular}
}
\caption{Our results and baselines on ScanNet-v2 val.~set. $^*$ means the baseline model trained with the initial pseudo labels shown in Figure~\ref{fig:overview} (d). $^\dagger$ means disabling graph propagation and relation network during inference, but note that they are still used in training. } 
\label{tab:baseline}
\end{table}

\vspace{-0.1in}

\paragraph{Results on ScanNet-v2 Data-Efficient Benchmark}
\label{sec:data-efficient}
In this section, we show results on ScanNet-v2 ``3D Semantic label with Limited Annotations'' benchmark~\cite{hou2020efficient}. We report the results on the most challenging setting with only 20 points annotated each scene in Table~\ref{tab:existing} ``Ours on Data Efficient'' and Table~\ref{tab:baseline} ``Data Efficient''. In this experiment, we use the officially provided 20 points instead of ``One-Thing-One-Click'', and then employ our self-training approach for semantic segmentation. 
The results show that our approach is not limited to ``One-Thing-One-Click'' and is applicable to other annotation schemes. However, the performance  is inferior to ``One-Thing-One-Click'', since the annotations are more uneven among the categories.

\vspace*{-0.15in}
\paragraph{Ablation Studies}

To further study the effectiveness of self-training, graph propagation and relation network, we conduct ablation studies on these three modules on ScanNet-v2 validation set as shown in Table~\ref{tab:ablation} with single view evaluation. 

``3D U-Net'' indicates that the labels are propagated only based on the confidence score of the 3D U-Net itself,~\ie, the unary term in Equation~\ref{equ:energy}. This ablation is designed to manifest the effectiveness of self-training. The ``3D U-Net'' column in Table~\ref{tab:ablation} manifests that the performance is consistently improved with self-training strategy even without pairwise energy term in Equation~\ref{equ:energy} and super-voxel partition. 

``3D U-Net+GP'' refers to the label propagation with graph model, and the similarity among super-voxels are measured by the coordinates $p_i$ and colors $c_i$ without the learned feature $f_i$. This ablation study is to show the effectiveness of the graph model. 
The results in Table~\ref{tab:ablation} indicate that the graph model benefits the label propagation, and finally boosts the overall performance by 2\% over ``3D U-Net'' (67.92\% vs. 65.91\%). 

``3D U-Net+Rel+GP'' utilizes the relation network for similarity measurement based on ``3D U-Net+GP''. In this setting, the similarity among super-voxels is measured with the averaged coordinates $p_i$, the colors $c_i$, the unary features $u_i$, and the relation network generated feature $f_i$, as shown in Equation~\ref{equ:energy}. This experiment is to manifest that the relation network benefits the similarity measurement and pseudo label generation, compared with the hand-crafted feature, i.e., coordinates and color. It outperforms the hand-crafted features especially in the later iterations since the network benefits from the richer pseudo labels. It finally achieves 2.5\% improvement compared with ``3D U-Net+GP'' (70.45\% vs. 67.92\%). 
As shown in Figure~\ref{fig:iter}, the generated pseudo labels for each iteration expands to unknown regions step by step and finally gets close to the ground truth.

\begin{table}
\centering
\scalebox{0.9}{
  \begin{tabular}{c|ccc}
    \toprule
    Method & 3D U-Net & 3D U-Net+GP &3D U-Net+Rel+GP \\
    \midrule
    Iter1 & 60.14  & 63.83 & \textbf{63.92}  \\
    Iter2 & 62.39  & 64.74 & \textbf{66.97} \\
    Iter3 & 64.83  & 66.10 & \textbf{68.40} \\
    Iter4 & 65.81  & 67.78 &  \textbf{70.01} \\
    Iter5 & 65.91  & 67.92 &  \textbf{70.45} \\
    \bottomrule
  \end{tabular}
}
\caption{Ablation studies. ``GP'' indicates the graph propagation, and ``Rel'' means the relation network. ``3D U-Net '' refers to propagating labels only with the network prediction itself.  ``3D U-Net+GP'' indicates label propagation with hand-crafted features. ``3D U-Net+Rel+GP'' indicates label propagation with our relation network. Evaluated on ScanNet-v2 val. set with single view testing. }
\label{tab:ablation}
\end{table}

\vspace*{-0.15in}

\paragraph{Analysis of Relation Network}

Further, we study whether the learned embeddings of the relation network outperform the hand-crafted features for similarity measurement. We randomly sample 200 super-voxels for each category in ScanNet-v2, and conduct a t-SNE visualization~\cite{maaten2008visualizing} on them. Figure~\ref{fig:tsne} indicates that the relation network better groups the intra-class embeddings and distinguish the inter-class embeddings compared with hand-crafted features.


\subsection{Evaluations on S3DIS}
We also evaluate our annotation and training approach on the S3DIS dataset. Only less than 0.02\% points in the dataset are annotated with our ``One Thing One Click'' scheme. To study whether the performance can be further boosted with richer annotations, we additionally conduct a ``One Thing Three Clicks'' scheme on S3DIS, where random 3 points per-object are annotated. 
\vspace*{-0.1in}
\paragraph{Comparing with Existing Works}

We also compare with fully supervised approaches and weakly supervised approaches on S3DIS. The latter includes existing works~\cite{laine2016temporal,tarvainen2017mean} and recent works~\cite{xu2020weakly,wang2020weakly}. 

As shown in Table~\ref{tab:s3dis}, with the ``One Thing One Click'' scheme where less than 0.02\% points are annotated, we achieve 50.1\% mIoU. With ``One Thing Three Clicks'' scheme, our performance can be further improved to 55.3\% mIoU. The above two results outperform~\cite{xu2020weakly} by 5.6\% and 10.8\% mIoU (0.2\% annotations in~\cite{xu2020weakly}), and 2.1\% and 7.3\% mIoU (10\% annotations in~\cite{xu2020weakly}) respectively. 

Wang~\etal~\cite{wang2020weakly} unprojects 2D semantic labels to 3D space for 3D semantic segmentation. To compare with~\cite{wang2020weakly}, we first compare with the actual number of annotated points regardless of 2D or 3D. For S3DIS, the number of annotated 2D pixels (70,496 images with 1080$\times$1080 resolution) is 100$\times$ more than the officially annotated 3D points ($5.27 \times 10^{8}$ in total), so both settings of~\cite{wang2020weakly} (100\% 2D annotations and 16.7\% 2D annotations) actually utilize a large quantity of annotations. 
Even with a large gap of annotation, the results in Table~\ref{tab:s3dis} show that our ``One Thing Three Clicks'' scheme with only 0.06\% 3D annotation outperforms~\cite{wang2020weakly} with 100\% 2D annotations by nearly 3\% mIoU.  

In addition, our approach achieves comparable results with several fully supervised methods as shown in Table~\ref{tab:s3dis}. 

\vspace*{-0.1in}

\paragraph{Comparing with Our Baselines }
We follow the similar settings in Section~\ref{sec:scannet} to show several baselines for S3DIS. 
\begin{itemize}
    \item Table~\ref{tab:s3dis} ``Our fully-sup baseline''. The model trained with the full supervision of S3DIS achieves 63.7\% mIoU. It serves as the upper bound of our approach. 
    \item The model directly trained with only the annotated points in Figure~\ref{fig:overview} (c) cannot converge well. 
    \item Table~\ref{tab:s3dis} ``One Thing One Click$^*$'' and ``One Thing Three Clicks$^*$''. The model trained with the annotated super-voxels in Figure~\ref{fig:overview} (d) achieves 43.7\% mIoU for ``One Thing One Click'' and 48.9\% mIoU for ``One Thing Three Clicks''. They are used as the baselines to calculate the ``relative improvement'' of our approach, and are denoted as ``our baseline'' in the following.
\end{itemize}

As shown in Table~\ref{tab:s3dis} ``Rel. Imp.'' column, we have 14.6\% (``One Thing One Click'') and 13.1\% (``One Thing Three Clicks'') relative improvement over our baseline, surpassing the relative improvement  of~\cite{xu2020weakly}, which is 1.1\% (with 0.2\% annotations) and 5\% (with 10\% annotations) over their own baselines, by a large margin. The significant improvement of ``relative improvement over baseline'' manifests the effectiveness of the proposed approach. 

To evaluate without any extra computation burden, we further disable the label propagation and relation network in inference as shown in Table~\ref{tab:s3dis} ``$^\dagger$''. Note that they are still adopted in training. Our model still attains 13.0\% (``One Thing One Click'') and 10.6\% (``One Thing Three Clicks'') relative improvement over our baseline in this case.  



\begin{table}
\centering
\scalebox{0.75}{
  \begin{tabular}{@{\hspace*{1mm}}c@{\hspace*{1mm}}|@{\hspace*{1mm}}c@{\hspace*{1mm}}|@{\hspace*{1mm}}c@{\hspace*{1mm}}|@{\hspace*{1mm}}c@{\hspace*{1mm}}}
    \toprule
    Method & Supervision (\%) & mIoU(\%) & Rel. Imp. (\%)\\
    \midrule
PointNet~\cite{qi2017pointnet}& 100\% & 41.1 &-\\
SegCloud~\cite{tchapmi2017segcloud}& 100\% &  48.9 &-\\
TangentConv~\cite{tatarchenko2018tangent}& 100\% & 52.8&-\\
3D RNN~\cite{ye20183d}& 100\% & 53.4 &-\\
PointCNN~\cite{li2018pointcnn}& 100\%& 57.3&-\\
SuperpointGraph~\cite{landrieu2018large}& 100\%& 58.0&-\\
MinkowskiNet32~\cite{choy20194d}& 100\% &65.4&-\\
Virtual MV-Fusion~\cite{kundu2020virtual} & 100\%+2D &65.4&-\\
    \midrule
    Our fully-sup baseline & 100\%& 63.7 &-\\
    \midrule
    $\mathbin{\Pi}$ Model~\cite{laine2016temporal} & 0.2\%&44.3&-\\
    MT~\cite{tarvainen2017mean}& 0.2\% & 44.4&-\\
    Xu~\etal
~\cite{xu2020weakly}$^*$ & 0.2\%  &  44.0&- \\
    Xu~\etal
~\cite{xu2020weakly} & 0.2\% &44.5&1.1 \\  
    $\mathbin{\Pi}$ Model~\cite{laine2016temporal} & 10\%&46.3&-\\
    MT~\cite{tarvainen2017mean} & 10\% &  47.9&-\\
   Xu~\etal~\cite{xu2020weakly}$^*$ & 10\% &45.7&-\\ 
    Xu~\etal~\cite{xu2020weakly} & 10\%  &48.0 &5.0\\    
    GPFN~\cite{wang2020weakly} & 16.7\% 2D& 50.8 &-\\  
    GPFN~\cite{wang2020weakly} & 100\% 2D& 52.5  &- \\
    \midrule
   One Thing One Click$^*$  & 0.02\% &43.7 & -\\
   One Thing One Click$^\dagger$ & 0.02\% & 49.4& \textbf{13.0}\\
   One Thing One Click & 0.02\% &\textbf{50.1} &\textbf{14.6} \\ 
    One Thing Three Clicks$^*$ & 0.06\%  & 48.9 &-\\
    One Thing Three Click$^\dagger$ & 0.06\% & 54.1&10.6\\
    One Thing Three Clicks & 0.06\%&\textbf{55.3}& 13.1 \\ 
    \bottomrule
  \end{tabular}
}
\caption{Comparison with existing methods and baselines on 
the S3DIS Area-5.
$^*$ indicates baseline models, and $^\dagger$ refers to disabling graph propagation and relation network during inference. Note that they are still used in training. ``Rel. Imp.'' indicates the relative improvement over the baseline. ``-'' indicates there is no meaningful baseline in this case or it is a baseline itself.}
\label{tab:s3dis}
\end{table}

\vspace*{-0.1in}
\section{Conclusion}

We propose the ``One Thing One Click'' scheme to efficiently annotate point clouds for weakly supervised 3D semantic segmentation, requiring significantly fewer annotations than the previous approaches.
%
To put this scheme into practice, we formulate a self-training approach to make it feasible for the network to learn from such extremely sparse labels.
Specifically, we execute the two key modules in our approach iteratively: expand labels through the graph propagation module and train the network using the updated pseudo labels.
Further, we adopt a relation network to explicitly learn the feature similarity among graph nodes with complex 3D structures. 
%
Experiments on two large 3D datasets ScanNet-v2 and S3DIS manifest that our approach, with only extremely-sparse annotations, outperforms all the existing weakly supervised methods on 3D semantic segmentation by a large margin, and our results are also comparable to those of the fully supervised counterparts.

\clearpage
\appendix
\centerline{\Large{\textbf{Supplementary Material}}}

In this supplementary document, we first show our results on ScanNet-v2 with fewer annotations (Section~\ref{sec:fewer}). Then, we show more results of using our method with ``One Thing One Click'' on ScanNet-v2 and S3DIS (Section~\ref{sec:illustration}).
Further, we discuss the relationship of our relation network and Prototypical Net~\cite{snell2017prototypical} (Section~\ref{sec:prototype}). 
Finally, we show example super-voxels in Section~\ref{sec:supervoxel}.
The code can be found in \url{https://github.com/liuzhengzhe/One-Thing-One-Click}.


\section{Results with Fewer Annotations}\label{sec:fewer}

To investigate the performance of our approach with even less annotated points, we further annotate ScanNet-v2 with a ``Two Things One Click'' scheme, where we annotate a single random point on half of the objects chosen randomly in the scene. In this way, only less than 0.01\% points are annotated on ScanNet-v2. With the even sparse annotations, we still achieve 60.62\% mIoU as shown in Table~\ref{tab:baseline}. This experiment also demonstrates that our method can still achieve decent performance even though the annotator ignores several objects by mistake in ``One Thing One Click'' scheme. We further investigate the performance drop with a more challenging ``Four Things One Click'' scheme. However, the model cannot converge well in the very first iteration due to the insufficient label and the self-training fails in this case.

\section{More Results on ScanNet-v2 and S3DIS}
\label{sec:illustration}

\begin{figure*}
\centering
\includegraphics[width=0.99\textwidth]{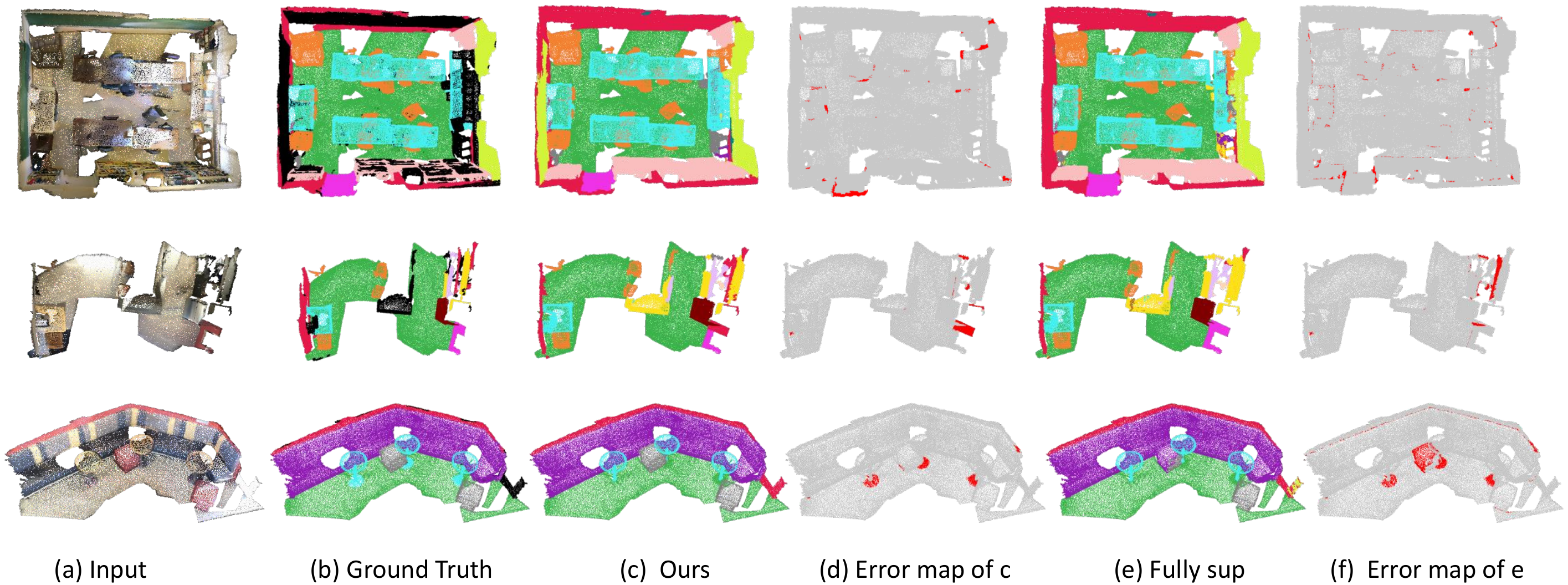}
\caption{More results on ScanNet-v2. (c) is produced by our model trained only with ``One Thing One Click'' annotations.
(e) is the fully supervised results of~\cite{graham2017submanifold}.
Red regions in (d) and (f) indicate the wrong predictions.
}
\label{fig:scannet_ill}
 \vspace*{-1mm}
\end{figure*}

\begin{figure*}
\centering
\includegraphics[width=0.99\textwidth]{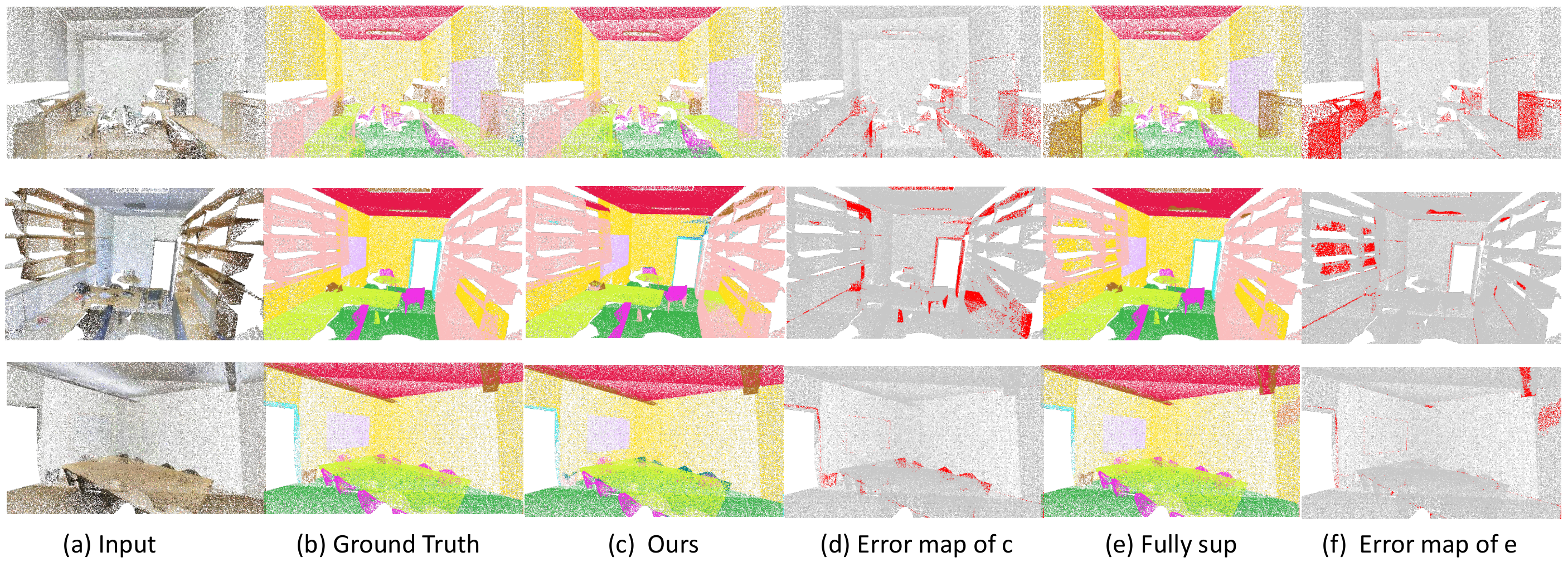}
\caption{More results on S3DIS. (c) is produced by our model trained only with ``One Thing One Click'' annotations.
(e) is the fully supervised results of~\cite{graham2017submanifold}.
Red regions in (d) and (f) indicate the wrong predictions.
}
\label{fig:s3dis_ill}
 \vspace*{-1mm}
\end{figure*}

In this section, we show more results on ScanNet-v2 and S3DIS in Figures~\ref{fig:scannet_ill} and~\ref{fig:s3dis_ill}, respectively.
Through these results, we demonstrate that even our approach is trained with only one annotated point per object in the scene, it can already produce segmentation results that are comparable to the fully supervised baseline~\cite{graham2017submanifold}.
See the error maps shown in (d) and (f) for better visualizations. 

\section{Relation to Prototypical Networks}\label{sec:prototype}

In this section, we discuss the relationship between our relation network and Prototypical Networks~\cite{snell2017prototypical}. First of all,~\cite{snell2017prototypical} focuses on few-shot learning, and requires a strong generalization ability to classify the categories that are not seen in the training.
In each training episode,~\cite{snell2017prototypical} samples a subset of categories to simulate the unseen categories in testing. For better simulation,~\cite{snell2017prototypical} does not require the prototype of each training category to be consistent in different training episodes. In this way, the network tends to regard the sampled training categories as unfamiliar, to better simulate the test case with new categories. Otherwise, the network will memorize the training categories themselves and lose the generalization ability to new categories in testing.
%

Very differently, in our method, the same categories are shared in both training and testing. To group the embedding of the same category and distinguish different categories, our categorical prototype should reveal the global mean representation of all samples in each category. To avoid the mean categorical embedding deviating from the actual categorical center, we design a memory bank in our model to update the prototypes with the moving average strategy, instead of relying on one single mini-batch.
Hence, we can stabilize the prototypes in the training and ensure that they are still effective in the inference. 

Secondly,~\cite{snell2017prototypical} focuses on the classification task and assumes that there are plenty of samples of each category in the training set to support the set construction in each episode. However, in our 3D semantic segmentation task, we sample point clouds in each iteration, and there could be insufficient or even no samples for certain categories in a mini-batch. For categories with insufficient samples, we sample them with replacement to match the number of samples of other categories. For categories with no samples in the batch, our memory bank helps to accumulate the embedding learned in the previous iterations and stabilize the prototypes of these categories relative to the actual categorical center.

Further, we conduct an ablation study to manifest the effectiveness of the memory bank in our relation network. In this ablation study (to be presented in the next section in this supplementary document), we adopt the same strategy as~\cite{snell2017prototypical} to update the prototype of each category. Specifically, we directly use the average embedding of the sampled super-voxels in the current mini-batch as the prototype, instead of updating the prototype using the memory bank. From the ablation study results presented in Table~\ref{tab:allablation}, ``3D U-Net+Rel (w/o MB)+GP'' shows that without memory bank, the performance of ``3D U-Net+Rel (w/o MB)+GP'' degrades to be a value similar to ``3D U-Net+GP,'' where the relation network is not used.  Please see Section~\ref{sec:allablation} for more details.

\begin{table}
\centering
\scalebox{0.9}{
  \begin{tabular}{c|ccc}
    \toprule
    Method & Annotation (\%) & mIoU (\%) \\
    \midrule
     Two Things One Click$^*$ & 0.01 & 54.71\\
     Two Things One Click$^\dagger$ & 0.01 & 59.56\\ 
     Two Things One Click & 0.01 & \textbf{60.62}\\ 
    \bottomrule
  \end{tabular}
}
\caption{Two Things One Click results and baselines on ScanNet-v2 val.~set. $^*$ means the baseline model trained with the initial pseudo label shown in Figure 3 (d). $^\dagger$ means disabling graph propagation and relation network during inference, but note that they are still used in training. } 
\label{tab:baseline}
\end{table}

\section{Summary of All the Ablation Studies}\label{sec:allablation}

\begin{table}[!t]
\centering
\scalebox{0.8}{
  \begin{tabular}{c|cc}
    \toprule
    Baselines & 3D U-Net (w/o ST) & 3D U-Net\\
    \midrule
    mIoU & 60.14 & 65.91 \\
    \toprule
    \toprule
    Baselines & 3D U-Net+GP & 3D U-Net+Rel (w/o MB)+GP   \\
    \midrule
    mIoU   & 67.92 & 67.98 \\
    \toprule
    \toprule
    Ours &  3D U-Net+Rel+GP$^\dagger$ & 3D U-Net+Rel+GP   \\
    \midrule
    mIoU & 69.12 &  \textbf{70.45}  \\
    \bottomrule
  \end{tabular}
}
\vspace{2mm}
\caption{Summary of our ablation studies.
``w/o'' is the short of ``without.''
``3D U-Net'' means the U-Net architecture in~\cite{graham2017submanifold}.
``ST'' means self-training,
``GP'' means graph propagation,
``Rel'' means relation network,
``MB'' means memory bank, and
``$^\dagger$'' indicates graph propagation and relation network are only utilized in training but not in inference.
From the results, we can see that adding components (ST, GP, etc.) in our method gradually improves the performance, and our full model achieves the best performance.}
\label{tab:allablation}
\end{table}

\begin{figure*}[!t]
\centering
\includegraphics[width=0.99\textwidth]{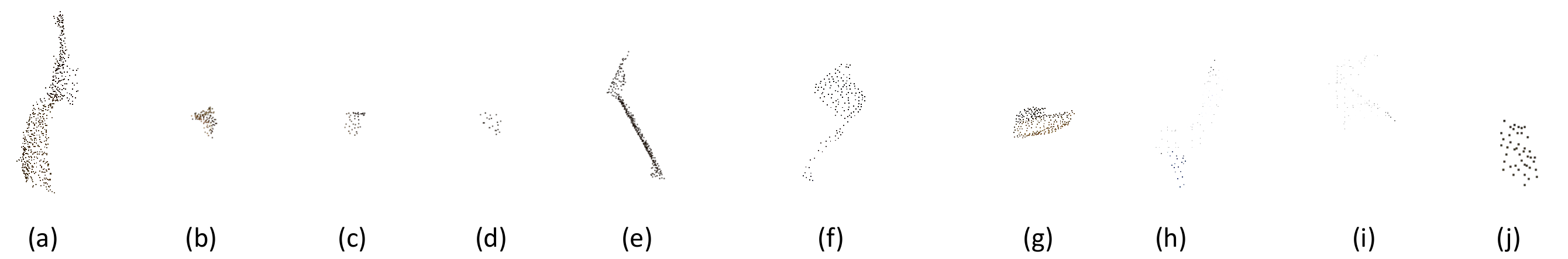}
\caption{Some example super-voxels in ScanNet-v2. They have large variations in shape, geometrical structure, density, and number of points.
}\label{fig:supervoxel}
 \vspace*{-1mm}
\end{figure*}

Table~\ref{tab:allablation} summarizes all the ablation studies presented in the main paper.
Their settings are listed below:
\begin{itemize}
\item 3D U-Net (w/o ST): Train~\cite{graham2017submanifold} with the ``Initial pseudo label'' as Figure 3(d) illustrated in the main paper, and without using our self training (ST) mechanism.
\item 3D U-Net: Self-training (ST) is adopted on ``3D U-Net (w/o ST).''
The pseudo labels for the second to fifth iterations are generated according to our self-training approach,~\ie, we use network predictions of high confidence to iteratively improve the results.
\item 3D U-Net+GP: Based on the previous model ``3D U-Net,'' we add back the graph propagation (GP) in the pseudo label generation, and utilize the hand-crafted features, including the colors and coordinates as the pairwise term for the similarity measurement.
\item 3D U-Net+Rel (w/o MB)+GP: Based on ``3D U-Net+GP,'' the pairwise term further includes the embedding generated from the relation network (Rel).
However, the categorical prototypes are derived as the mean of the sampled data in the current mini-batch without using the memory bank (MB).
This setting is discussed earlier in Section~\ref{sec:prototype} of this supplementary document but not included in the main paper.
\item 3D U-Net+Rel+GP$^\dagger$: Based on ``3D U-Net+Rel (w/o MB)+GP,'' we further utilize the memory bank (MB) to update the categorical prototypes.
However, graph propagation and relation network are used only in the training but not in the inference.
This is to evaluate the performance of our approach with the same computation complexity as ``3D U-Net'' in the inference.
\item 3D U-Net+Rel+GP: This is our full model, for which its only difference compared to ``3D U-Net+Rel+GP$^\dagger$'' is that it utilizes graph propagation and relation network in both training and inference.
\end{itemize}

From Table~\ref{tab:allablation}, we can see that each of the key modules of our approach, i.e., self-training (ST), graph propagation (GP), relation network (Rel), and memory bank (MB), has its own contribution to the overall performance.

\section{Examples of the Super-Voxel}\label{sec:supervoxel}

Lastly, we show example super-voxels in ScanNet-v2 in Figure~\ref{fig:supervoxel}. Due to the irregular geometrical structures and complex shapes, hand-crafted features like colors and coordinates cannot fully describe their properties. To this end, we propose a relation network to learn the high-level similarities among them.

{\small
\bibliographystyle{ieee_fullname}
\bibliography{cvpr}

\begin{thebibliography}{10}\itemsep=-1pt

\bibitem{ahn2018learning}
Jiwoon Ahn and Suha Kwak.
\newblock Learning pixel-level semantic affinity with image-level supervision
  for weakly supervised semantic segmentation.
\newblock In {\em Proceedings of the IEEE Conference on Computer Vision and
  Pattern Recognition (CVPR)}, pages 4981--4990, 2018.

\bibitem{armeni2017joint}
Iro Armeni, Sasha Sax, Amir~R. Zamir, and Silvio Savarese.
\newblock Joint 2{D}-3{D}-semantic data for indoor scene understanding.
\newblock {\em arXiv preprint arXiv:1702.01105}, 2017.

\bibitem{bearman2016s}
Amy Bearman, Olga Russakovsky, Vittorio Ferrari, and Li Fei-Fei.
\newblock What’s the point: Semantic segmentation with point supervision.
\newblock In {\em European Conference on Computer Vision (ECCV)}, pages
  549--565. Springer, 2016.

\bibitem{boulch2020convpoint}
Alexandre Boulch.
\newblock Conv{P}oint: Continuous convolutions for point cloud processing.
\newblock {\em Computers \& Graphics}, 2020.

\bibitem{chen2017deeplab}
Liang-Chieh Chen, George Papandreou, Iasonas Kokkinos, Kevin Murphy, and
  Alan~L. Yuille.
\newblock Deep{L}ab: Semantic image segmentation with deep convolutional nets,
  atrous convolution, and fully connected {CRF}s.
\newblock {\em IEEE Transactions on Pattern Analysis and Machine Intelligence
  (T-PAMI)}, 40(4):834--848, 2017.

\bibitem{choy20194d}
Christopher Choy, JunYoung Gwak, and Silvio Savarese.
\newblock 4{D} spatio-temporal {ConvNets}: Minkowski convolutional neural
  networks.
\newblock In {\em Proceedings of the IEEE Conference on Computer Vision and
  Pattern Recognition (CVPR)}, pages 3075--3084, 2019.

\bibitem{dai2017scannet}
Angela Dai, Angel~X. Chang, Manolis Savva, Maciej Halber, Thomas Funkhouser,
  and Matthias Nie{\ss}ner.
\newblock Scan{N}et: Richly-annotated {3D} reconstructions of indoor scenes.
\newblock In {\em Proc. Computer Vision and Pattern Recognition (CVPR), IEEE},
  2017.

\bibitem{dai20183dmv}
Angela Dai and Matthias Nie{\ss}ner.
\newblock 3{DMV}: Joint 3{D}-multi-view prediction for 3{D} semantic scene
  segmentation.
\newblock In {\em Proceedings of the European Conference on Computer Vision
  (ECCV)}, pages 452--468, 2018.

\bibitem{dai2015boxsup}
Jifeng Dai, Kaiming He, and Jian Sun.
\newblock Box{S}up: Exploiting bounding boxes to supervise convolutional
  networks for semantic segmentation.
\newblock In {\em Proceedings of the IEEE International Conference on Computer
  Vision (CVPR)}, pages 1635--1643, 2015.

\bibitem{dong2018tri}
Wei~Gao Dong-Dong~Chen, Wei~Wang and Zhi~Hua Zhou.
\newblock Tri-net for semi-supervised deep learning.
\newblock In {\em International Joint Conferences on Artificial Intelligence
  (IJCAI)}, 2018.

\bibitem{graham2015sparse}
Ben Graham.
\newblock Sparse 3{D} convolutional neural networks.
\newblock {\em arXiv preprint arXiv:1505.02890}, 2015.

\bibitem{graham2017submanifold}
Benjamin Graham, Martin Engelcke, and Laurens van~der Maaten.
\newblock 3{D} semantic segmentation with submanifold sparse convolutional
  networks.
\newblock {\em Proceedings of the IEEE Conference on Computer Vision and
  Pattern Recognition (CVPR)}, 2018.

\bibitem{guinard2017weakly}
St{\'e}phane Guinard and Loic Landrieu.
\newblock Weakly supervised segmentation-aided classification of urban scenes
  from 3{D} lidar point clouds.
\newblock In {\em ISPRS Workshop 2017}, 2017.

\bibitem{han2020occuseg}
Lei Han, Tian Zheng, Lan Xu, and Lu Fang.
\newblock Occuseg: Occupancy-aware 3{D} instance segmentation.
\newblock In {\em Proceedings of the IEEE/CVF Conference on Computer Vision and
  Pattern Recognition (CVPR)}, pages 2940--2949, 2020.

\bibitem{he2020momentum}
Kaiming He, Haoqi Fan, Yuxin Wu, Saining Xie, and Ross Girshick.
\newblock Momentum contrast for unsupervised visual representation learning.
\newblock In {\em Proceedings of the IEEE/CVF Conference on Computer Vision and
  Pattern Recognition (CVPR)}, pages 9729--9738, 2020.

\bibitem{hou2020efficient}
Ji Hou, Benjamin Graham, Matthias Nie{\ss}ner, and Saining Xie.
\newblock Exploring data-efficient 3d scene understanding with contrastive
  scene contexts.
\newblock {\em arXiv preprint arXiv:2012.09165}, 2020.

\bibitem{hu2020jsenet}
Zeyu Hu, Mingmin Zhen, Xuyang Bai, Hongbo Fu, and Chiew-lan Tai.
\newblock {JSEN}et: Joint semantic segmentation and edge detection network for
  3{D} point clouds.
\newblock {\em arXiv preprint arXiv:2007.06888}, 2020.

\bibitem{jiang2020pointgroup}
Li Jiang, Hengshuang Zhao, Shaoshuai Shi, Shu Liu, Chi-Wing Fu, and Jiaya Jia.
\newblock Pointgroup: Dual-set point grouping for 3{D} instance segmentation.
\newblock In {\em Proceedings of the IEEE/CVF Conference on Computer Vision and
  Pattern Recognition (CVPR)}, pages 4867--4876, 2020.

\bibitem{koller2009probabilistic}
Daphne Koller and Nir Friedman.
\newblock {\em Probabilistic graphical models: principles and techniques}.
\newblock MIT press, 2009.

\bibitem{kundu2020virtual}
Abhijit Kundu, Xiaoqi~Michael Yin, Alireza Fathi, David~Alexander Ross, Brian
  Brewington, Tom Funkhouser, and Caroline Pantofaru.
\newblock Virtual multi-view fusion for {3D} semantic segmentation.
\newblock In {\em Proceedings of the European Conference on Computer Vision
  (ECCV)}, 2020.

\bibitem{laine2016temporal}
Samuli Laine and Timo Aila.
\newblock Temporal ensembling for semi-supervised learning.
\newblock {\em arXiv preprint arXiv:1610.02242}, 2016.

\bibitem{landrieu2018large}
Loic Landrieu and Martin Simonovsky.
\newblock Large-scale point cloud semantic segmentation with superpoint graphs.
\newblock In {\em Proceedings of the IEEE Conference on Computer Vision and
  Pattern Recognition (CVPR)}, pages 4558--4567, 2018.

\bibitem{laradji2018blobs}
Issam~H Laradji, Negar Rostamzadeh, Pedro~O Pinheiro, David Vazquez, and Mark
  Schmidt.
\newblock Where are the blobs: Counting by localization with point supervision.
\newblock In {\em Proceedings of the European Conference on Computer Vision
  (ECCV)}, pages 547--562, 2018.

\bibitem{li2018pointcnn}
Yangyan Li, Rui Bu, Mingchao Sun, Wei Wu, Xinhan Di, and Baoquan Chen.
\newblock Point{CNN}: Convolution on x-transformed points.
\newblock In {\em Advances in neural information processing systems (NeurIPS)},
  pages 820--830, 2018.

\bibitem{lin2016scribblesup}
Di Lin, Jifeng Dai, Jiaya Jia, Kaiming He, and Jian Sun.
\newblock Scribble{S}up: Scribble-supervised convolutional networks for
  semantic segmentation.
\newblock In {\em Proceedings of the IEEE Conference on Computer Vision and
  Pattern Recognition (CVPR)}, pages 3159--3167, 2016.

\bibitem{lin2020fpconv}
Yiqun Lin, Zizheng Yan, Haibin Huang, Dong Du, Ligang Liu, Shuguang Cui, and
  Xiaoguang Han.
\newblock F{PC}onv: Learning local flattening for point convolution.
\newblock In {\em Proceedings of the IEEE/CVF Conference on Computer Vision and
  Pattern Recognition (CVPR)}, pages 4293--4302, 2020.

\bibitem{maaten2008visualizing}
Laurens van~der Maaten and Geoffrey Hinton.
\newblock Visualizing data using t-{SNE}.
\newblock {\em Journal of machine learning research}, 9(Nov):2579--2605, 2008.

\bibitem{maninis2018deep}
Kevis-Kokitsi Maninis, Sergi Caelles, Jordi Pont-Tuset, and Luc Van~Gool.
\newblock Deep extreme cut: From extreme points to object segmentation.
\newblock In {\em Proceedings of the IEEE Conference on Computer Vision and
  Pattern Recognition (CVPR)}, pages 616--625, 2018.

\bibitem{mei2019semantic}
Jilin Mei, Biao Gao, Donghao Xu, Wen Yao, Xijun Zhao, and Huijing Zhao.
\newblock Semantic segmentation of 3{D} lidar data in dynamic scene using
  semi-supervised learning.
\newblock {\em IEEE Transactions on Intelligent Transportation Systems},
  21(6):2496--2509, 2019.

\bibitem{oh2017exploiting}
Seong~Joon Oh, Rodrigo Benenson, Anna Khoreva, Zeynep Akata, Mario Fritz, and
  Bernt Schiele.
\newblock Exploiting saliency for object segmentation from image level labels.
\newblock In {\em 2017 IEEE conference on computer vision and pattern
  recognition (CVPR)}, pages 5038--5047. IEEE, 2017.

\bibitem{oord2018representation}
Aaron van~den Oord, Yazhe Li, and Oriol Vinyals.
\newblock Representation learning with contrastive predictive coding.
\newblock {\em arXiv preprint arXiv:1807.03748}, 2018.

\bibitem{papadopoulos2017extreme}
Dim~P Papadopoulos, Jasper~RR Uijlings, Frank Keller, and Vittorio Ferrari.
\newblock Extreme clicking for efficient object annotation.
\newblock In {\em Proceedings of the IEEE international conference on computer
  vision (ICCV)}, pages 4930--4939, 2017.

\bibitem{NEURIPS2019_9015}
Adam Paszke, Sam Gross, Francisco Massa, Adam Lerer, James Bradbury, Gregory
  Chanan, Trevor Killeen, Zeming Lin, Natalia Gimelshein, Luca Antiga, Alban
  Desmaison, Andreas Kopf, Edward Yang, Zachary DeVito, Martin Raison, Alykhan
  Tejani, Sasank Chilamkurthy, Benoit Steiner, Lu Fang, Junjie Bai, and Soumith
  Chintala.
\newblock Py{T}orch: An imperative style, high-performance deep learning
  library.
\newblock In H. Wallach, H. Larochelle, A. Beygelzimer, F. d\textquotesingle
  Alch\'{e}-Buc, E. Fox, and R. Garnett, editors, {\em Advances in Neural
  Information Processing Systems (NeurIPS) 32}, pages 8024--8035. Curran
  Associates, Inc., 2019.

\bibitem{qi2017pointnet}
Charles~R. Qi, Hao Su, Kaichun Mo, and Leonidas~J Guibas.
\newblock Point{N}et: Deep learning on point sets for 3{D} classification and
  segmentation.
\newblock In {\em Proceedings of the IEEE conference on computer vision and
  pattern recognition (CVPR)}, pages 652--660, 2017.

\bibitem{qi2017pointnet++}
Charles~Ruizhongtai Qi, Li Yi, Hao Su, and Leonidas~J Guibas.
\newblock Point{N}et++: Deep hierarchical feature learning on point sets in a
  metric space.
\newblock In {\em Advances in neural information processing systems (NeurIPS)},
  pages 5099--5108, 2017.

\bibitem{qi2016augmented}
Xiaojuan Qi, Zhengzhe Liu, Jianping Shi, Hengshuang Zhao, and Jiaya Jia.
\newblock Augmented feedback in semantic segmentation under image level
  supervision.
\newblock In {\em European conference on computer vision (ECCV)}, pages
  90--105. Springer, 2016.

\bibitem{qiao2018deep}
Siyuan Qiao, Wei Shen, Zhishuai Zhang, Bo Wang, and Alan Yuille.
\newblock Deep co-training for semi-supervised image recognition.
\newblock In {\em Proceedings of the European Conference on Computer Vision
  (ECCV)}, pages 135--152, 2018.

\bibitem{riegler2017octnet}
Gernot Riegler, Ali Osman~Ulusoy, and Andreas Geiger.
\newblock Octnet: Learning deep 3{D} representations at high resolutions.
\newblock In {\em Proceedings of the IEEE Conference on Computer Vision and
  Pattern Recognition (CVPR)}, pages 3577--3586, 2017.

\bibitem{schult2020dualconvmesh}
Jonas Schult, Francis Engelmann, Theodora Kontogianni, and Bastian Leibe.
\newblock {DualConvMesh-Net}: Joint geodesic and euclidean convolutions on 3{D}
  meshes.
\newblock In {\em Proceedings of the IEEE/CVF Conference on Computer Vision and
  Pattern Recognition (CVPR)}, pages 8612--8622, 2020.

\bibitem{snell2017prototypical}
Jake Snell, Kevin Swersky, and Richard Zemel.
\newblock Prototypical networks for few-shot learning.
\newblock In {\em Advances in neural information processing systems (NeurIPS)},
  pages 4077--4087, 2017.

\bibitem{su2018splatnet}
Hang Su, Varun Jampani, Deqing Sun, Subhransu Maji, Evangelos Kalogerakis,
  Ming-Hsuan Yang, and Jan Kautz.
\newblock Splat{N}et: Sparse lattice networks for point cloud processing.
\newblock In {\em Proceedings of the IEEE Conference on Computer Vision and
  Pattern Recognition (CVPR)}, pages 2530--2539, 2018.

\bibitem{tarvainen2017mean}
Antti Tarvainen and Harri Valpola.
\newblock Mean teachers are better role models: Weight-averaged consistency
  targets improve semi-supervised deep learning results.
\newblock In {\em Advances in neural information processing systems (NeurIPS)},
  pages 1195--1204, 2017.

\bibitem{tatarchenko2018tangent}
Maxim Tatarchenko, Jaesik Park, Vladlen Koltun, and Qian-Yi Zhou.
\newblock Tangent convolutions for dense prediction in 3{D}.
\newblock In {\em Proceedings of the IEEE Conference on Computer Vision and
  Pattern Recognition (CVPR)}, pages 3887--3896, 2018.

\bibitem{tchapmi2017segcloud}
Lyne Tchapmi, Christopher Choy, Iro Armeni, JunYoung Gwak, and Silvio Savarese.
\newblock Segcloud: Semantic segmentation of 3{D} point clouds.
\newblock In {\em 2017 international conference on 3D vision (3DV)}, pages
  537--547. IEEE, 2017.

\bibitem{thomas2019kpconv}
Hugues Thomas, Charles~R. Qi, Jean-Emmanuel Deschaud, Beatriz Marcotegui,
  Fran{\c{c}}ois Goulette, and Leonidas~J Guibas.
\newblock Kp{C}onv: Flexible and deformable convolution for point clouds.
\newblock In {\em Proceedings of the IEEE International Conference on Computer
  Vision (ICCV)}, pages 6411--6420, 2019.

\bibitem{wang2019boundary}
Bin Wang, Guojun Qi, Sheng Tang, Tianzhu Zhang, Yunchao Wei, Linghui Li, and
  Yongdong Zhang.
\newblock Boundary perception guidance: A scribble-supervised semantic
  segmentation approach.
\newblock In {\em International Joint Conference on Artificial Intelligence
  (IJCAI)}, pages 3663--3669, 2019.

\bibitem{wang2020weakly}
Haiyan Wang, Xuejian Rong, Liang Yang, Jinglun Feng, Jizhong Xiao, and Yingli
  Tian.
\newblock Weakly supervised semantic segmentation in 3{D} graph-structured
  point clouds of wild scenes.
\newblock {\em arXiv preprint arXiv:2004.12498}, 2020.

\bibitem{wei2020multi}
Jiacheng Wei, Guosheng Lin, Kim-Hui Yap, Tzu-Yi Hung, and Lihua Xie.
\newblock Multi-path region mining for weakly supervised 3{D} semantic
  segmentation on point clouds.
\newblock In {\em Proceedings of the IEEE/CVF Conference on Computer Vision and
  Pattern Recognition (CVPR)}, pages 4384--4393, 2020.

\bibitem{wu2019pointconv}
Wenxuan Wu, Zhongang Qi, and Li Fuxin.
\newblock Point{C}onv: Deep convolutional networks on 3{D} point clouds.
\newblock In {\em Proceedings of the IEEE Conference on Computer Vision and
  Pattern Recognition (CVPR)}, pages 9621--9630, 2019.

\bibitem{wu2018unsupervised}
Zhirong Wu, Yuanjun Xiong, Stella~X Yu, and Dahua Lin.
\newblock Unsupervised feature learning via non-parametric instance
  discrimination.
\newblock In {\em Proceedings of the IEEE Conference on Computer Vision and
  Pattern Recognition (CVPR)}, pages 3733--3742, 2018.

\bibitem{xie2020pointcontrast}
Saining Xie, Jiatao Gu, Demi Guo, Charles~R Qi, Leonidas Guibas, and Or Litany.
\newblock Pointcontrast: Unsupervised pre-training for 3d point cloud
  understanding.
\newblock In {\em European Conference on Computer Vision}, pages 574--591.
  Springer, 2020.

\bibitem{xu2020weakly}
Xun Xu and Gim~Hee Lee.
\newblock Weakly supervised semantic point cloud segmentation: Towards 10x
  fewer labels.
\newblock In {\em Proceedings of the IEEE/CVF Conference on Computer Vision and
  Pattern Recognition (CVPR)}, pages 13706--13715, 2020.

\bibitem{ye20183d}
Xiaoqing Ye, Jiamao Li, Hexiao Huang, Liang Du, and Xiaolin Zhang.
\newblock 3{D} recurrent neural networks with context fusion for point cloud
  semantic segmentation.
\newblock In {\em Proceedings of the European Conference on Computer Vision
  (ECCV)}, pages 403--417, 2018.

\bibitem{yuan2019structpool}
Hao Yuan and Shuiwang Ji.
\newblock Struct{P}ool: Structured graph pooling via conditional random fields.
\newblock In {\em International Conference on Learning Representations (ICLR)},
  2019.

\bibitem{zhang2020weakly}
Jing Zhang, Xin Yu, Aixuan Li, Peipei Song, Bowen Liu, and Yuchao Dai.
\newblock Weakly-supervised salient object detection via scribble annotations.
\newblock In {\em Proceedings of the IEEE/CVF Conference on Computer Vision and
  Pattern Recognition (CVPR)}, pages 12546--12555, 2020.

\bibitem{zheng2015conditional}
Shuai Zheng, Sadeep Jayasumana, Bernardino Romera-Paredes, Vibhav Vineet,
  Zhizhong Su, Dalong Du, Chang Huang, and Philip~HS Torr.
\newblock Conditional random fields as recurrent neural networks.
\newblock In {\em Proceedings of the IEEE international conference on computer
  vision (ICCV)}, pages 1529--1537, 2015.

\bibitem{zhou2018weakly}
Yanzhao Zhou, Yi Zhu, Qixiang Ye, Qiang Qiu, and Jianbin Jiao.
\newblock Weakly supervised instance segmentation using class peak response.
\newblock In {\em Proceedings of the IEEE Conference on Computer Vision and
  Pattern Recognition (CVPR)}, pages 3791--3800, 2018.

\end{thebibliography}
}

\end{document}